\documentclass{article}

\PassOptionsToPackage{numbers, compress}{natbib}

    \usepackage[preprint]{neurips_2020}

\usepackage[utf8]{inputenc} 
\usepackage[T1]{fontenc}    
\usepackage{hyperref}       
\usepackage{url}            
\usepackage{booktabs}       
\usepackage{amsfonts}       
\usepackage{nicefrac}  
\usepackage{graphicx}
\usepackage{amsmath}
\usepackage{amssymb}
\usepackage{subfigure}
\usepackage{microtype}      
\usepackage{amsmath} 
\usepackage[ruled,vlined]{algorithm2e}
\usepackage{setspace}
\usepackage{graphicx}

\DeclareMathOperator*{\argmaxA}{arg\,max}
\DeclareMathOperator*{\maxA}{max}

\title{Robustness to Augmentations as a Generalization Metric}

\author{
  Sumukh Aithal K \thanks{Equal Contribution} \\
  Department of Computer Science\\
  PES University\\
  Bengaluru, India \\
  \texttt{sumukhaithal6@gmail.com} \\
  
   \And
   Dhruva Kashyap \footnotemark[1] \\
   Department of Computer Science\\
  PES University\\
  Bengaluru, India \\
   \texttt{dhruva12kashyap@gmail.com} \\
   \AND
   Natarajan Subramanyam \\
  Department of Computer Science\\
  PES University\\
  Bengaluru, India \\
  \texttt{natarajan@pes.edu} \\
}

\begin{document}

\maketitle

\begin{abstract}
Generalization is the ability of a model to predict on unseen domains and is a fundamental task in machine learning. Several generalization bounds, both theoretical and empirical have been proposed but they do not provide tight bounds. In this work, we propose a simple yet effective method to predict the generalization performance of a model by using the concept that models that are robust to augmentations are more generalizable than those which are not. We experiment with several augmentations and composition of augmentations to check the generalization capacity of a model. We also provide a detailed motivation behind the proposed method. The proposed generalization metric is calculated based on the change in the model's output after augmenting the input.
The proposed method was the first runner up solution for the competition "Predicting Generalization in Deep Learning".
\end{abstract}


\section{Introduction}
Deep learning models are more of a black box technique and it is hard to explain why a model behaves in the way it does. Despite this limitation, there is the tremendous success of deep neural networks in a wide variety of tasks ranging from image classification to speech recognition \cite{graves2013speech,krizhevsky2017imagenet}. But there is still a need to able to understand why these models work so well.
\\
Generalization is one of the fundamental problems in machine learning and estimating the generalization performance of a model is not trivial.  Many generalization metrics have been proposed but they tend to overestimate the generalization performance. Recent developments in the field are focused on the calculation of a complexity measure, which is a quantity to directly indicate the extent of generalizability of a model. There have been theoretical \cite{arora2018stronger,neyshabur2015norm} and empirical \cite{keskar2016large} measures that can assess the degree of generalization of a model. Theoretical complexity measures such as VC-dimension \cite{JMLR:v20:17-612} has been the standard for measuring generalizability as there has been an established monotonic relation between the VC-dimension and the generalizability of a model.
\\
Generalization performance is very crucial in practical computer vision and models need to adapt and generalize well to unseen domains. Current computer vision models don’t generalize very well to unseen domains and there has been extensive work in domain adaptation \cite{ganin2015unsupervised} and domain generalization to fit new domains.  
\\
The generalization gap of a model is defined as the difference between the estimated risk of a target function and the empirical risk of a target function. The estimated risk is not computable but is estimated as the risk of the target function on a validation set and the empirical risk is estimated as the risk of the target function on the training set. The task of the competition was to predict the generalization of a model through a complexity measure that maps the model and dataset to a real number. This real number indicates the generalization ability of the model. 

\section{Related Work}
\citet{Jiang*2020Fantastic} presents a large scale study of various generalization metrics that have been applied in deep neural networks. They explore 40 complexity measures, both theoretical and experimental, and provide metrics for the evaluation of generalization bounds. 
They discuss conducting a conditional independent test by calculating the conditional mutual information between the generalization metric and the observed generalization in a causal probabilistic graph. This causality is conditioned over a set of hyper-parameters of the model. 
\\
\citet{keskar2016large} suggested using sharpness as a generalization measure and sharpness measures the robustness to adversarial perturbations. They also showed the success of small batch methods over large batch methods when trained with Stochastic Gradient Descent and showed that training with large batch methods tends to converge to sharp minima which leads to poorer generalization.
\\
\citet{neyshabur2017exploring} investigates different complexity measures like norm, PAC-Bayes to explain the generalization of deep networks. They also show that combining sharpness with PAC-Bayes analysis with the norm of the weights.
\citet{jiang2018predicting} proposed to use the margin distribution as a predictor for the generalization gap in deep neural networks.

\citet{dziugaite2017computing} proposes direct optimization on PAC-Bayes bounds to compute non-vacuous numerical bounds on generalization error in deep neural networks with more model parameters than training data.   
\citet{kawaguchi2017generalization} has discussed theoretical motivations for the ability of deep neural networks to generalize, and provides new open problems in the field. 

\section{Method}
\subsection{Motivation}

Many deep neural networks have used data augmentation to prevent overfitting. But recent work \cite{azulay2019deep}, shows convolutional neural networks are very sensitive to small geometric transformations to the input that are imperceptible to humans even when the model is trained on a wide range of augmented data. 
\\
The proposed method is based on a simple hypothesis that a model capable of generalizing must be robust to augmentations. 
A model’s output should not change significantly when certain augmentations are performed on the input. 
In other words, a model’s predicted class for input should not change when we augment the input without erasing key features from the input. 
For example, if a picture of a cat which the model correctly classified is augmented with a rotation by 180 degrees, then the model must still predict it as a cat. The model should confidently predict the augmented input if it has learned the correct features of a particular class. 
\\
Humans predominantly classify objects based on their shape rather than texture whereas most CNNs classify based on texture rather than shape \cite{geirhos2018imagenettrained}. Augmentations that perform changes to the texture would be a suitable test of a model’s generalizability.

The generalization power can also be looked at from the point of view of the model’s performance on input from a shifted distribution. Based on this, we augment images in such a way that important features of the image are retained.

\subsection{Proposed Metric}
Algorithm \ref{mainalgo} describes the proposed generalization metric.
\begin{algorithm}[ht]
\SetAlgoLined
\textbf{Input: }Consider a model $\theta$;  x is the input; $\lambda$ is the penalty for an augmentation.\\
\KwResult{Generalization metric \(\phi\) }
 \(\phi\) = 0 \;
 \ForAll{samples x}{
  x' = Augment(x) \;
  \eIf{$\argmaxA_{\hat y} P_\theta(\hat y|x) = \argmaxA_{\hat y} P_\theta(\hat y | x')$}{
     $\phi = \phi \ - \mid\maxA_{\hat y} P_{\theta}(\hat y|x) - P_{\theta}(\hat y | x')\mid \;$ 
  }{
    $\phi = \phi - \lambda \; $
  }}
 \caption{Proposed metric calculation}
 \label{mainalgo}
\end{algorithm}
For every sample, we augment the input and then compare the class prediction of the model for the original input and augmented input. If the class prediction is the same even after the input is augmented, then we add a penalty equal to the difference between probabilities of the predicted class on the original and augmented input. This represents the difference in the confidence in the class when the input is augmented. The penalty is determined based on the strength of the augmentation. 
\\
The strength of augmentation is determined by the ability of the augmentation to change the texture in the input. Augmentations that do not alter the texture of the image, but to tend to alter the shape in the image, are weak. 
\\
For example, flipping an image left to right does not change any texture in the input, only the shape. Augmentations that alter the texture in an image are referred to as strong augmentations. These augmentations may or may not alter the shape of the image, but they affect the texture in the image. For example, applying a Sobel filter \cite{kanopoulos1988design} on an image removes most of the texture in the image and retains only an edge map.
\\
A model that misclassifies samples on weak augmentations are considered to generalize poorly are penalized heavily. Models that misclassify samples with strong augmentations are penalized less.
The generalization metric reflects the penalty that has been incurred over an augmented subset of the training data. Thus, models that accumulate a large penalty have a higher negative score and tend to generalize poorly.

\begin{figure*}[ht]
  \centering
  \subfigure[Original Image ]{\includegraphics[width=0.3\linewidth]{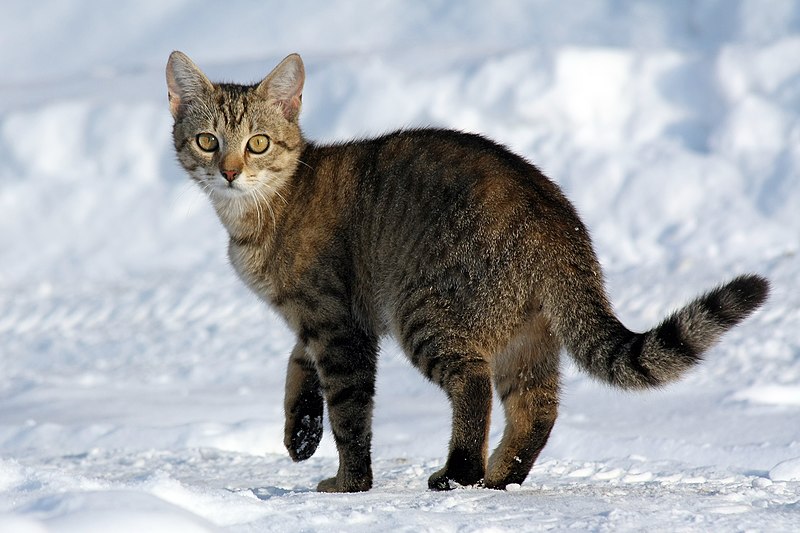}}\quad
  \subfigure[Center Crop]{\includegraphics[width=0.3\linewidth]{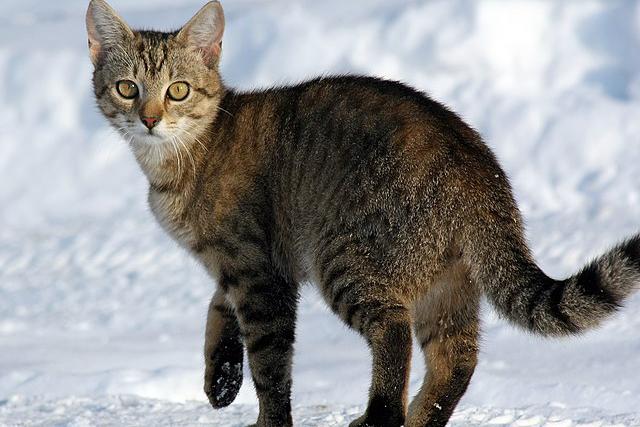}}\quad
  \subfigure[Flip Left Right]{\includegraphics[width=0.3\linewidth]{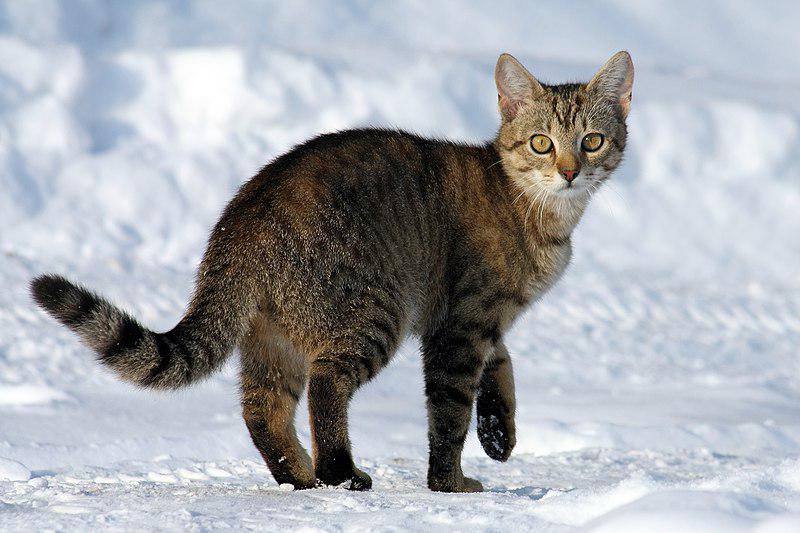}}\quad
    \\
  \subfigure[Random Saturation ]{\includegraphics[width=0.3\linewidth]{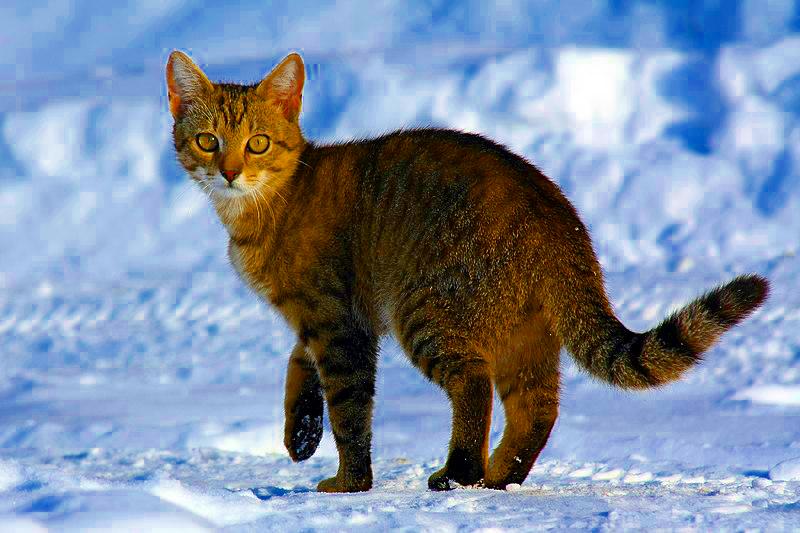}}\quad
  \subfigure[Random Erasing]{\includegraphics[width=0.3\linewidth]{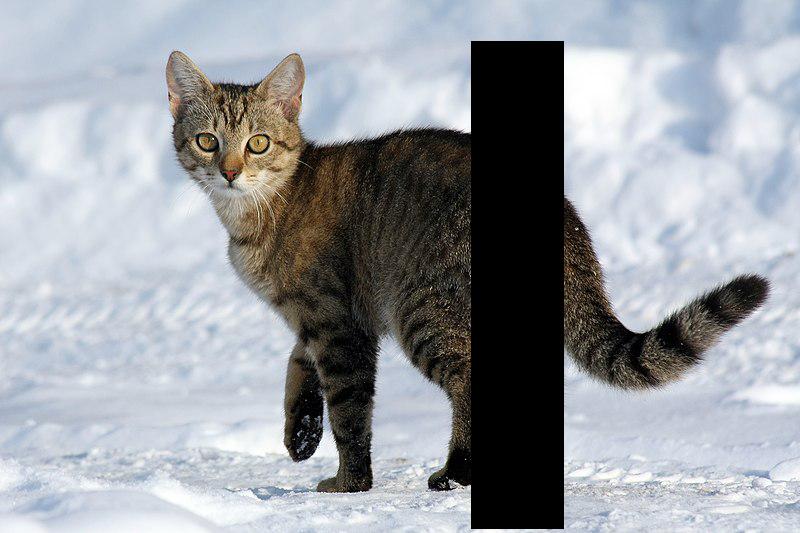}}\quad
  \subfigure[Sobel Filter]{\includegraphics[width=0.3\linewidth]{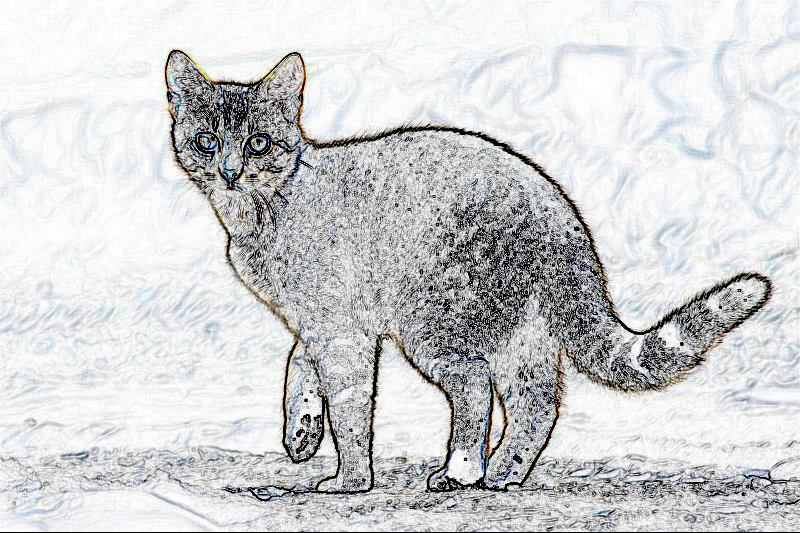}}\quad
  \caption{Illustration of experimented augmentations. (Original image cc-by: Von.grzanka)}
  \label{fig:images}
\end{figure*}
\subsection{List of Augmentations}
\begin{enumerate}
    \item Flip: Flips the image on the vertical axis
    \item Random Saturation: Randomly increases the saturation of each pixel
    \item Crop and Resize: Crops a central portion of the image and resizes it to the original dimensions
    \item Brightness: Increases brightness of each pixel
    \item Random Erasing \cite{zhong1708random}: Erases a random grid of the image
    \item Sobel Filter \cite{kanopoulos1988design}: Provides an edge map of the image
    \item Virtual Adversarial Perturbation\cite{miyato2018virtual}: Inspired by the metric of sharpness \cite{keskar2016large}, a similar method which measures the robustness of the model against local perturbation was used as an augmentation.
\end{enumerate}

Figure \ref{fig:images} is a visual depiction of some of the augmentations. 
We experimented with individual and composition of augmentations like flip and then add saturation, which yielded good results.
We also experimented with neural style transfer \cite{gatys2015neural,geirhos2018imagenettrained} which makes the image texture-invariant, but were unable to produce good results probably because of the texture bias of most of the models.
\section{Results}
These are the results on the public dataset calculated on the evaluation metric, Conditional Mutual Information \cite{Jiang*2020Fantastic}.
The public dataset of this competition involves two groups of models trained on two publicly available datasets, namely, CIFAR-10 \cite{krizhevsky2009learning} and SVHN \cite{netzer2011reading}. The models were trained on different model architectures and different training schemes. The models trained on CIFAR-10 had a VGG like architecture trained with different batch sizes, learning rate, dropout rate, convolution layers, and dense layers. The training datasets consisted of a total of 150 models trained on different architectures with different hyperparameters. Higher the score, the better the generalization metric.

\begin{table}
\begin{center}
\resizebox{\textwidth}{!}
{\begin{tabular}{ c c c c c c c c c c }
\toprule $\lambda_{flip}$ &  $\lambda_{saturation}$ & $\lambda_{crop\_resize}$ & $\lambda_{sobel} $ & $\lambda_{brightness}$ & $\lambda_{flip+saturation}$ & $\lambda_{cutout}$ & $\lambda_{v}$ &  Public Score & Private Score \\
\hline
6 & 1 & 3 & 2 & 1 & 12 & 0 & 3 & 33.67 & 9.16 \\
6 & 1 & 2 & 3 & 1 & 9 & 0 & 0 & 40.9  & 9.25 \\
6 & 1 & 2 & 3 & 1 & 12 & 2 & 0 & \textbf{41.8} & \textbf{10.6} \\
\bottomrule
\end{tabular}}
\end{center}
\caption{Penalties for misclassification on each augmentation and scores obtained on 3 submissions}
\label{penalties}
\end{table}
Table \ref{penalties} shows the penalties for each of the augmentations and the respective scores on both the public and private datasets. $\lambda$ indicates the penalty value for the respective augmentation and $\lambda_{v}$ indicates the penalty for the virtual adversarial perturbation. It can be seen that a high penalty value for the composition of simple augmentations seem to work well in the datasets and using stronger augmentations like cutout seem to have a positive effect on the score.

We had experimented with each of these augmentations individually, although we were able to score good results, we found that compounding each augmentation with relative penalties provided greater scores. This ensemble of augmentations allowed us to measure the robustness of a model against various strong and weak augmentations. The penalties for each augmentation were empirically calculated. 

\section{Conclusion}

Calculation of the generalization bound on a deep learning network is a critical field of study and extensive work is being done in the field. We have provided a measure of a model's ability to generalize based on conditional mutual information. This method tests the model with augmented samples from the training data and penalizes models that are unable to correctly classify augmented samples. We have also provided an insight into the role of augmentations in testing the generalizability of a model. This metric can be used to assess the generalization capacity of any deep learning model.

\bibliography{sample}

\begin{thebibliography}{20}
\providecommand{\natexlab}[1]{#1}
\providecommand{\url}[1]{\texttt{#1}}
\expandafter\ifx\csname urlstyle\endcsname\relax
  \providecommand{\doi}[1]{doi: #1}\else
  \providecommand{\doi}{doi: \begingroup \urlstyle{rm}\Url}\fi

\bibitem[Arora et~al.(2018)Arora, Ge, Neyshabur, and Zhang]{arora2018stronger}
S.~Arora, R.~Ge, B.~Neyshabur, and Y.~Zhang.
\newblock Stronger generalization bounds for deep nets via a compression
  approach.
\newblock \emph{arXiv preprint arXiv:1802.05296}, 2018.

\bibitem[Azulay and Weiss(2019)]{azulay2019deep}
A.~Azulay and Y.~Weiss.
\newblock Why do deep convolutional networks generalize so poorly to small
  image transformations?, 2019.

\bibitem[Bartlett et~al.(2019)Bartlett, Harvey, Liaw, and
  Mehrabian]{JMLR:v20:17-612}
P.~L. Bartlett, N.~Harvey, C.~Liaw, and A.~Mehrabian.
\newblock Nearly-tight vc-dimension and pseudodimension bounds for piecewise
  linear neural networks.
\newblock \emph{Journal of Machine Learning Research}, 20\penalty0
  (63):\penalty0 1--17, 2019.
\newblock URL \url{http://jmlr.org/papers/v20/17-612.html}.

\bibitem[Dziugaite and Roy(2017)]{dziugaite2017computing}
G.~K. Dziugaite and D.~M. Roy.
\newblock Computing nonvacuous generalization bounds for deep (stochastic)
  neural networks with many more parameters than training data.
\newblock \emph{arXiv preprint arXiv:1703.11008}, 2017.

\bibitem[Ganin and Lempitsky(2015)]{ganin2015unsupervised}
Y.~Ganin and V.~Lempitsky.
\newblock Unsupervised domain adaptation by backpropagation.
\newblock In \emph{International conference on machine learning}, pages
  1180--1189. PMLR, 2015.

\bibitem[Gatys et~al.(2015)Gatys, Ecker, Bethge, and Sep]{gatys2015neural}
L.~A. Gatys, A.~S. Ecker, M.~Bethge, and C.~Sep.
\newblock A neural algorithm of artistic style. arxiv 2015.
\newblock \emph{arXiv preprint arXiv:1508.06576}, 2015.

\bibitem[Geirhos et~al.(2019)Geirhos, Rubisch, Michaelis, Bethge, Wichmann, and
  Brendel]{geirhos2018imagenettrained}
R.~Geirhos, P.~Rubisch, C.~Michaelis, M.~Bethge, F.~A. Wichmann, and
  W.~Brendel.
\newblock Imagenet-trained {CNN}s are biased towards texture; increasing shape
  bias improves accuracy and robustness.
\newblock In \emph{International Conference on Learning Representations}, 2019.
\newblock URL \url{https://openreview.net/forum?id=Bygh9j09KX}.

\bibitem[Graves et~al.(2013)Graves, Mohamed, and Hinton]{graves2013speech}
A.~Graves, A.-r. Mohamed, and G.~Hinton.
\newblock Speech recognition with deep recurrent neural networks.
\newblock In \emph{2013 IEEE international conference on acoustics, speech and
  signal processing}, pages 6645--6649. IEEE, 2013.

\bibitem[Jiang et~al.(2019)Jiang, Krishnan, Mobahi, and
  Bengio]{jiang2018predicting}
Y.~Jiang, D.~Krishnan, H.~Mobahi, and S.~Bengio.
\newblock Predicting the generalization gap in deep networks with margin
  distributions.
\newblock In \emph{International Conference on Learning Representations}, 2019.
\newblock URL \url{https://openreview.net/forum?id=HJlQfnCqKX}.

\bibitem[Jiang* et~al.(2020)Jiang*, Neyshabur*, Mobahi, Krishnan, and
  Bengio]{Jiang*2020Fantastic}
Y.~Jiang*, B.~Neyshabur*, H.~Mobahi, D.~Krishnan, and S.~Bengio.
\newblock Fantastic generalization measures and where to find them.
\newblock In \emph{International Conference on Learning Representations}, 2020.
\newblock URL \url{https://openreview.net/forum?id=SJgIPJBFvH}.

\bibitem[Kanopoulos et~al.(1988)Kanopoulos, Vasanthavada, and
  Baker]{kanopoulos1988design}
N.~Kanopoulos, N.~Vasanthavada, and R.~L. Baker.
\newblock Design of an image edge detection filter using the sobel operator.
\newblock \emph{IEEE Journal of solid-state circuits}, 23\penalty0
  (2):\penalty0 358--367, 1988.

\bibitem[Kawaguchi et~al.(2017)Kawaguchi, Kaelbling, and
  Bengio]{kawaguchi2017generalization}
K.~Kawaguchi, L.~P. Kaelbling, and Y.~Bengio.
\newblock Generalization in deep learning.
\newblock \emph{arXiv preprint arXiv:1710.05468}, 2017.

\bibitem[Keskar et~al.(2016)Keskar, Mudigere, Nocedal, Smelyanskiy, and
  Tang]{keskar2016large}
N.~S. Keskar, D.~Mudigere, J.~Nocedal, M.~Smelyanskiy, and P.~T.~P. Tang.
\newblock On large-batch training for deep learning: Generalization gap and
  sharp minima.
\newblock \emph{arXiv preprint arXiv:1609.04836}, 2016.

\bibitem[Krizhevsky et~al.(2009)Krizhevsky, Hinton,
  et~al.]{krizhevsky2009learning}
A.~Krizhevsky, G.~Hinton, et~al.
\newblock Learning multiple layers of features from tiny images.
\newblock 2009.

\bibitem[Krizhevsky et~al.(2017)Krizhevsky, Sutskever, and
  Hinton]{krizhevsky2017imagenet}
A.~Krizhevsky, I.~Sutskever, and G.~E. Hinton.
\newblock Imagenet classification with deep convolutional neural networks.
\newblock \emph{Communications of the ACM}, 60\penalty0 (6):\penalty0 84--90,
  2017.

\bibitem[Miyato et~al.(2018)Miyato, Maeda, Koyama, and
  Ishii]{miyato2018virtual}
T.~Miyato, S.-i. Maeda, M.~Koyama, and S.~Ishii.
\newblock Virtual adversarial training: a regularization method for supervised
  and semi-supervised learning.
\newblock \emph{IEEE transactions on pattern analysis and machine
  intelligence}, 41\penalty0 (8):\penalty0 1979--1993, 2018.

\bibitem[Netzer et~al.(2011)Netzer, Wang, Coates, Bissacco, Wu, and
  Ng]{netzer2011reading}
Y.~Netzer, T.~Wang, A.~Coates, A.~Bissacco, B.~Wu, and A.~Y. Ng.
\newblock Reading digits in natural images with unsupervised feature learning.
\newblock 2011.

\bibitem[Neyshabur et~al.(2015)Neyshabur, Tomioka, and
  Srebro]{neyshabur2015norm}
B.~Neyshabur, R.~Tomioka, and N.~Srebro.
\newblock Norm-based capacity control in neural networks.
\newblock In \emph{Conference on Learning Theory}, pages 1376--1401, 2015.

\bibitem[Neyshabur et~al.(2017)Neyshabur, Bhojanapalli, McAllester, and
  Srebro]{neyshabur2017exploring}
B.~Neyshabur, S.~Bhojanapalli, D.~McAllester, and N.~Srebro.
\newblock Exploring generalization in deep learning.
\newblock In \emph{Advances in neural information processing systems}, pages
  5947--5956, 2017.

\bibitem[Zhong et~al.()Zhong, Zheng, Kang, Li, and Yang]{zhong1708random}
Z.~Zhong, L.~Zheng, G.~Kang, S.~Li, and Y.~Yang.
\newblock Random erasing data augmentation. arxiv 2017.
\newblock \emph{arXiv preprint arXiv:1708.04896}.

\end{thebibliography}

\end{document}